\pdfoutput=1

\documentclass[11pt]{article}
\usepackage[]{emnlp2021}

\usepackage{subcaption}
\usepackage{times}
\usepackage{latexsym}
\usepackage{amsmath}
\usepackage{mathrsfs}
\usepackage{graphicx}
\usepackage{booktabs}       
\usepackage{float}

\usepackage[T1]{fontenc}

\usepackage[utf8]{inputenc}

\usepackage{microtype}

\title{Zero-Shot Learning in Named-Entity Recognition \\ with External Knowledge}

\author{
    Nguyen Van Hoang
    \and Soeren Hougaard Mulvad
    \and Dexter Neo Yuan Rong
    \and Yang Yue \\
    National University of Singapore / School of Computing\\
    \texttt{\{vhnguyen,shmulvad,e0534450,yue.yang\}@u.nus.edu}
}

\begin{document}
\maketitle
\begin{abstract}
A significant shortcoming of current state-of-the-art (SOTA) named-entity recognition (NER) systems is their lack of generalization to unseen domains, which poses a major problem since obtaining labeled data for NER in a new domain is expensive and time-consuming.
We propose ZERO, a model that performs zero-shot and few-shot learning in NER to generalize to unseen domains by incorporating pre-existing knowledge in the form of semantic word embeddings. ZERO first obtains contextualized word representations of input sentences using the model LUKE, reduces their dimensionality, and compares them directly with the embeddings of the external knowledge, allowing ZERO to be trained to recognize unseen output entities.
We find that ZERO performs well on unseen NER domains with an average macro F1 score of 0.23, outperforms LUKE in few-shot learning, and even achieves competitive scores on an in-domain comparison. The performance across source-target domain pairs is shown to be inversely correlated with the pairs' KL divergence.
\end{abstract}

\section{Introduction}

Named-entity recognition (NER) is the common NLP task of classifying named entities appearing in texts into pre-defined categories. The most prevalent type of NER is in-domain recognition, in which the model is trained on a well-studied domain that contains a large amount of labeled data such as news and also evaluated on texts from the same domain.

However, it is common that texts might originate from different domains and we need to perform NER well on a new domain \cite{lee2018transfer}. But retraining the model on the target domain is not necessarily an option since labeled data of good quality can be difficult, expensive, and time-consuming to obtain. Label scarcity is especially pronounced for patient note de-identification, the task of removing protected health information from medical notes. Tuning NER systems to perform well in a new domain by hand is possible, but requires significant effort and is often not feasible \cite{lee2018transfer}. In this scenario, we have to resort to cross-domain NER. Having a system that can perform well on domains where labeled data is scarce or even non-existing would be extremely valuable.

We propose ZERO, a model that  performs zero-shot or few-shot learning in NER with external knowledge. The external knowledge comes in the form of word embeddings originating from either GloVe or Conceptnet Numberbatch \cite{sennington2014glove,speer2019ensemble,speer2017conceptnet}.

We obtain word embeddings for each label as well as entity representations from LUKE (Language Understanding with Knowledge-based Embeddings) for each token in the sentences \cite{yamada2020luke}. For each entity, we calculate dot products for all possible entity-label pairs and obtain the corresponding probabilities for these pairs. The label with the highest probability will be the output label. This enables us to obtain a probability value directly from the pre-trained embeddings rather than relying on having learned the labels during training.

The main contributions of our work are:
\begin{itemize}
    \item ZERO would allow training a well-performing NER model without any labeled examples in target domains which saves efforts and expenses on labeling.
    \item ZERO outperforms LUKE in few-shot learning. When there are only a few samples in the target domain, ZERO serves as a better choice to accomplish the NER tasks.
    \item We find that the KL divergence is inversely correlated with ZERO's performance and can be used to estimate how well it will perform on a given source-target domain transfer.
\end{itemize}

\section{Related work}

\subsection{LUKE}
We use pre-trained LUKE (Language Understanding with Knowledge-based Embeddings) as our backbone architecture \cite{yamada2020luke}. LUKE is based on a transformer trained on a large amount of entity-annotated corpora obtained from Wikipedia \cite{vaswani2017attention}. A key difference between LUKE and other contextualized word representations (CWRs) is that LUKE regards both words and entities as independent tokens allowing intermediate and output representations to be computed for all tokens using the transformer (see Figure \ref{fig:zero}). Since entities are regarded as tokens, LUKE can be used to directly model the relationships between entities.

\subsection{Zero-Shot Learning}
Zero-shot learning is the task of predicting labels not seen during training. Pioneering research on zero-shot learning began in the area of computer vision and used human-engineered features encoded as vectors. These vectors denote the presence or absence of any of the features \cite{5206772,5206594}. The model then learns a direct mapping between the image features and the class vectors and is able to predict unseen class encodings.

\subsection{Commonsense Knowledge and Word Embeddings}
Commonsense knowledge is the type of knowledge that most humans are expected to know about the everyday world, e.g.\ ``a car is capable of driving''. The current SOTA commonsense knowledge graph is ConceptNet which employs knowledge from a combination of crowdsourced and expert-created resources \cite{speer2017conceptnet}.

Roy et al.\ describe an approach for encoding commonsense knowledge to a zero-shot learning system using ConceptNet and a graph convolution network-based autoencoder \cite{roy2020common}. However, for NLP-related tasks where we can directly encode a ``meaning'' in the words in the form of word embeddings, this step can be avoided. Word2Vec popularized the concept of word embeddings and was since improved upon by GloVe \cite{mikolov2013word2vec, sennington2014glove}. The Conceptnet Numberbatch is an ensemble of the word embeddings Word2Vec, GloVe, and FastText retrofitted with the ConceptNet knowledge graph \cite{bojanowski2017fasttext, speer2019ensemble}. This retrofitted embedding is shown to be superior to e.g.\ GloVe for a wide array of tasks \cite{speer2019ensemble}.

\section{Methodology}

\begin{figure*}[ht]
    \begin{center}
    \includegraphics[width=\textwidth]{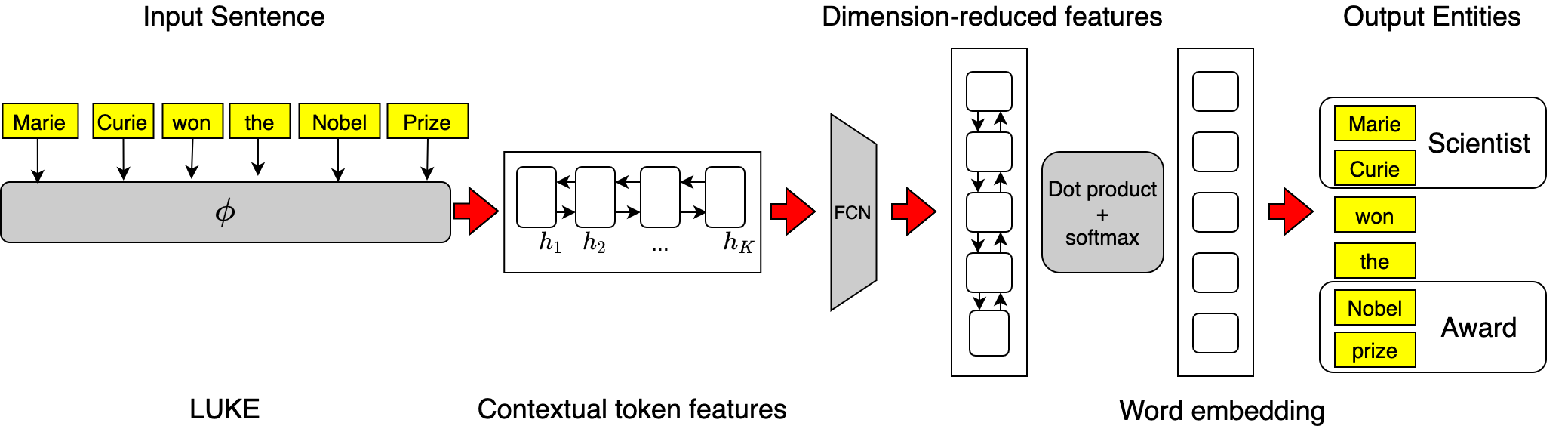}
    \end{center}
    \caption{ZERO - Zero-shot learning architecture.}
    \label{fig:zero}
\end{figure*}

The architecture of ZERO is shown in Figure \ref{fig:zero}. The first input of the model is a tokenized sentence. Each token is assumed to have an embedding representation $\mathbf{x}_i$, $i \in 1, \dots, K$ where $K$ equals the number of tokens in the input sentence. The second input is the source and target domain, which we use to indicate the corresponding label set.

Inspired by the most recent works on NER, we adopt a network base to obtain a contextualized representation of words and entities \cite{devlin2019bert}.
For this component, which we indicate as $\phi$, we take inspiration from the LUKE architecture since it is the current state-of-the-art of traditional NER \cite{yamada2020luke}. LUKE outputs contextualized representation for each word and entity.
The LUKE model is trained on a set of downstream tasks, such as NER and Q\&A.
Since the LUKE network has no knowledge of what the domain of the input sentence is, we use the output representation as the contextual token features.
We denote such a vector as $\mathbf{h}_i=\phi(\mathbf{x}_i)$, where $i \in 1, \dots, K$.

Once given the domain, each label entity features $\mathbf{e}_j$ are derived from the pre-trained distributed word embedding, such as GloVe or Conceptnet Numberbatch that aim to capture the label's semantics.

An FCN then reduces the dimensionality of $\mathbf{h}_i$ to a more compact representation before we finally compute the dot product of each token feature $\mathbf{h}_i$ with each of the label entity features $\mathbf{e}_j$. We use Eq.\ (\ref{eq:zero-eq}) to annotate the $i$-th token with the final prediction.
\begin{equation} \label{eq:zero-eq}
    y_i = \underset{j}{\mathrm{argmax}}(\mathrm{softmax}(\mathbf{h}_i \cdot \mathbf{e}_j))
\end{equation}

\section{Experiments}

\subsection{Dataset}
We use the CrossNER dataset \cite{liu2020crossner}. It consists of a general news domain \textit{Reuters} as well as the five domains \textit{AI}, \textit{literature} (LI), \textit{music} (MU), \textit{politics} (PO), and \textit{natural science} (NS) with 900-1400 labeled NER samples per domain. The domains have shared general categories such as ``person'' and ``location'' and their own specialized entity categories such as ``book'' and ``poem'' for the domain of literature.

\subsection{Results}

\begin{figure}[ht!]
    \begin{center}
    \includegraphics[width=\linewidth]{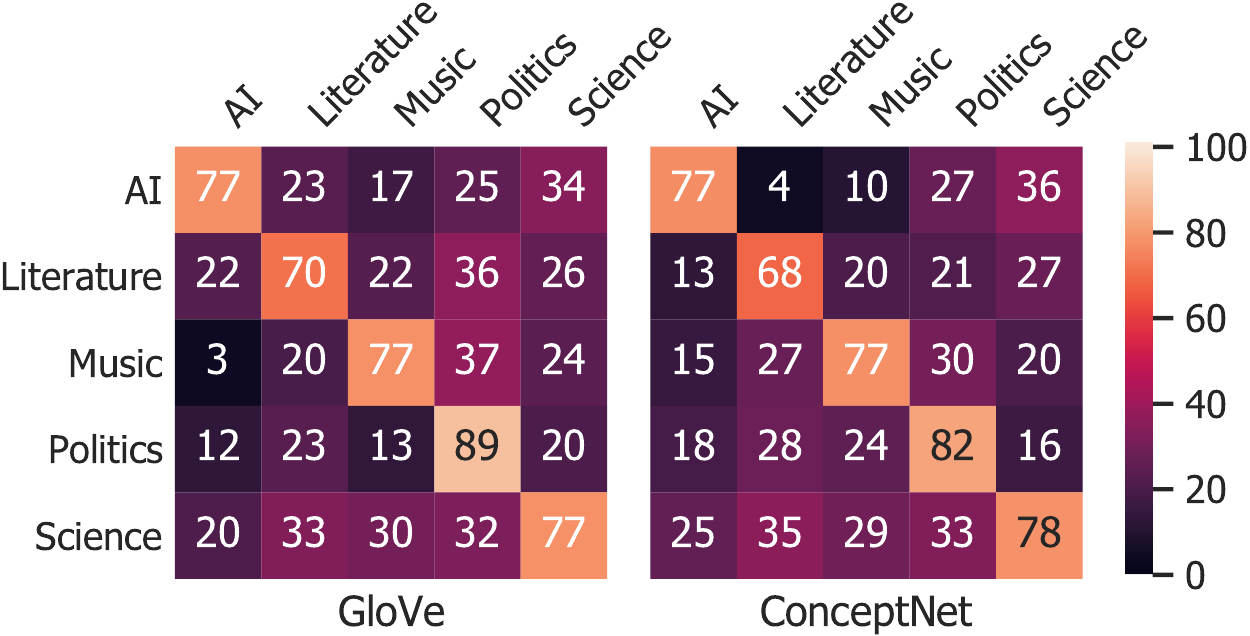}
    \end{center}
    \caption{Macro F1 scores for zero-shot learning (multiplied by 100 for readability). Rows are the source (train) domains and columns are the target (test) domains. The average of non-diagonal values is 0.2323 (GloVe) and 0.2283 (ConceptNet).
    }
    \label{fig:word-embed-res}
\end{figure}

\begin{figure}[ht!]
    \begin{center}
    \includegraphics[width=\linewidth]{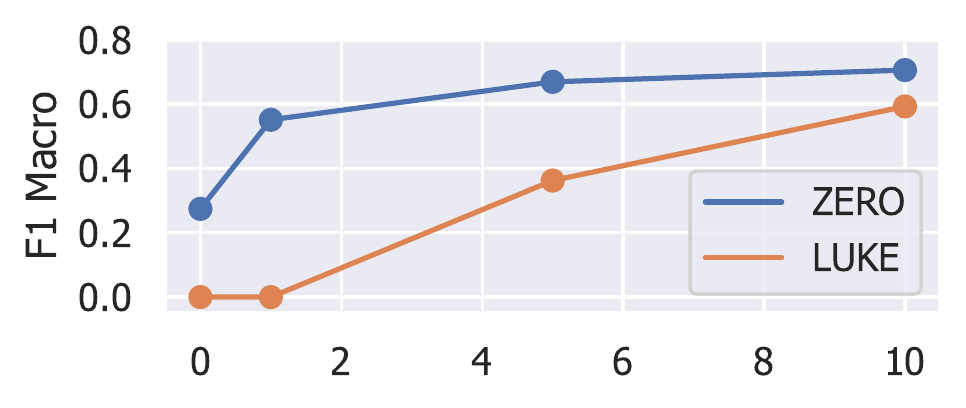}
    \end{center}
    \caption{Macro F1 scores for few-shot learning. $n \in \{0, 1, 5, 10\}$ training examples per label from the target domain seen during training. Scores are the average of the domain pairs of music-science and science-music.}
    \label{fig:few-shot-learning}
\end{figure}

The results for a zero-shot learning, few-shot learning, and in-domain context are shown in respectively Figure \ref{fig:word-embed-res}, Figure \ref{fig:few-shot-learning}, and Table \ref{tab:in-domain}.

\begin{table}[ht!]
\centering
\begin{tabular}{lccccc}
\hline
  & AI     & LI & MU  & PO & NS \\ \hline
Z & 0.769          & 0.703          & 0.772          & \textbf{0.886} & 0.770 \\
L & \textbf{0.772} & \textbf{0.747} & \textbf{0.871} & 0.875          & \textbf{0.783} \\ \hline
\end{tabular}
\caption{Macro F1 scores in an in-domain context of respectively ZERO (Z) using GloVe and LUKE (L). In-domain refers to the source and target domain being the same.}
\label{tab:in-domain}
\end{table}

\section{Discussions}
\subsection{Word Embedding Comparison}

In Figure \ref{fig:word-embed-res}, we observe noticeable better macro F1 scores of ZERO when we use GloVe for the underlying word embeddings as compared to Conceptnet Numberbatch. This is an interesting finding, since as mentioned Conceptnet Numberbatch is shown to be superior to GloVe for a wide array of tasks.

We hypothesized that while Conceptnet is superior in general, GloVe might be particularly good for the specific labels and concepts in CrossNER. To test our hypothesis using a quantifiable method, we extract the embeddings for all labels and concepts. With these, we perform unsupervised clustering using $k$-Means where we know in reality there should be 5 clusters. Comparing the obtained clusters to the ground truth labels with standard clustering performance metrics in Table \ref{tab:cluter-metrics}, we find that GloVe indeed results in better clusters for our task.

\begin{table}[H]
\centering
\begin{tabular}{lrr}
\hline
           & Adj.\ Rand Score & V-Measure       \\ \hline
GloVe      & \textbf{0.0630}     & \textbf{0.1230} \\
ConceptNet & -0.0307             & 0.0971          \\ \hline
\end{tabular}
\caption{Adjusted rand score and V-Measure when performing $k$-Means with $k=5$.}
\label{tab:cluter-metrics}
\end{table}

\subsection{Correlation Between Domain Similarity and Performance}

We explore whether the performance obtained on all domain pairs using the GloVe embeddings is correlated with the domain similarity. Vocabulary distributions are established for each domain based on word frequencies of the corpora of these domains and then all distributions are defined on the same probability space. We calculate the KL divergence between all domain pairs $P-Q$ as in Eq.\ (\ref{eq:kl}) to represent domain distances. $D_{\textrm{KL}} (P \ || \ Q)$, the KL divergence from domain $Q$ to domain $P$, is regarded as the distance when the model is transferred from domain $Q$ to domain $P$. For all words $w \in \mathscr{V}$ where $\mathscr{V}$ is the global vocabulary space, we denote $P(w)$ as the probability distribution function for $w$ in domain $P$.
\begin{equation} \label{eq:kl}
    D_{\textrm{KL}} (P\ || \ Q) = \sum_{w\in\mathscr{V}} P(w)\log \left( \frac{P(w)}{Q(w)} \right)
\end{equation}

\begin{figure}[H]
    \begin{center}
    \includegraphics[width=\linewidth]{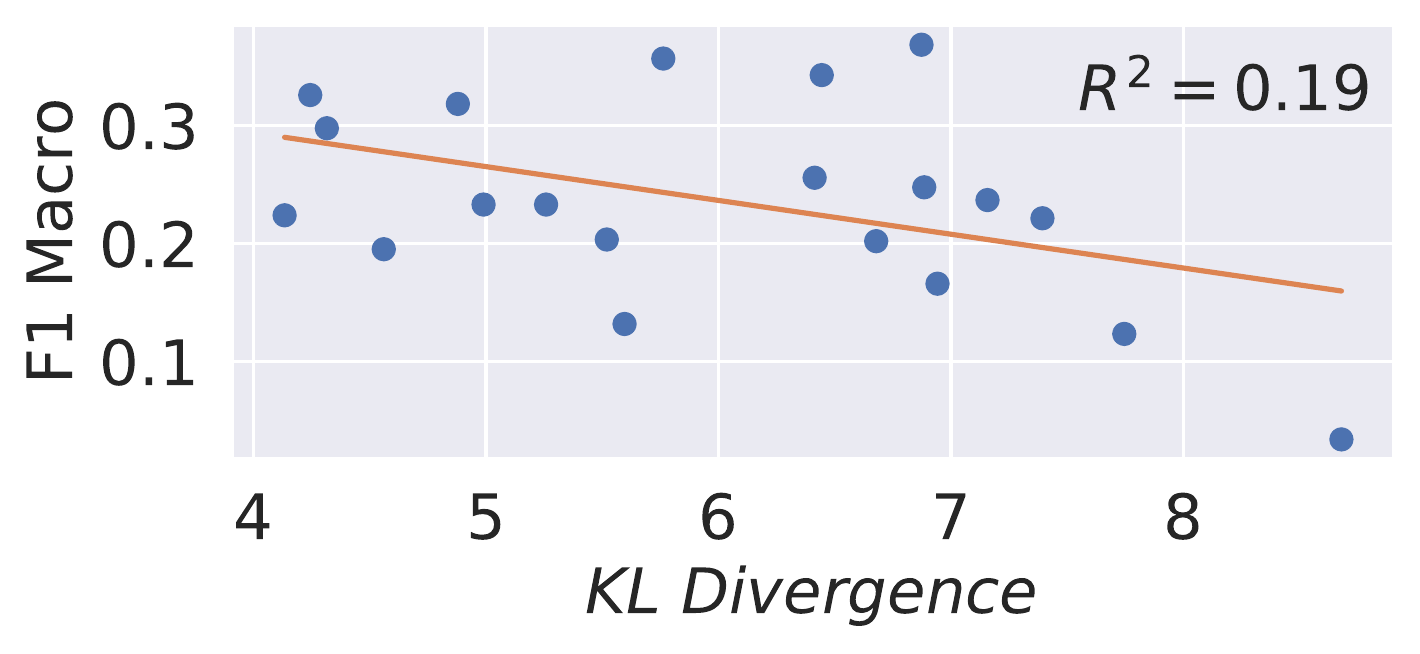}
    \end{center}
    \caption{Correlation between domain similarity (divergence) and performance.}
    \label{fig:per-sim}
\end{figure}

A regression line is fit based on the performances and distances. The regression result is displayed in Figure \ref{fig:per-sim} which shows an inverse correlation between the F1 macro score and KL divergence with an $R^2$ of $0.19$. This gives us an intuition that if the similarity from one domain to another domain is very high the corresponding domain adaptation will have a higher tendency to perform well.

\subsection{In-Domain Comparison With LUKE}

We evaluate the strength of our approach by comparing the in-domain performance of ZERO against LUKE, meaning that both models are trained and tested in the same domain. As shown in Table \ref{tab:in-domain}, we find that jointly learning with additional domain features is able to achieve competitive macro F1 scores. Except for the music domain, ZERO only performs slightly worse than LUKE in the domains of AI, literature, and science.

\subsection{Few-Shot Learning}

We examine how effective ZERO is in a few-shot learning setting. We add a few training examples per label from the target domain and compare the performance of ZERO in terms of F1 score against LUKE where both are trained on the original training examples from the source domain and the additional few training examples from the target domain. Our observations show that ZERO significantly outperforms LUKE when the number of training examples on the target domain is small. Specifically, given a single training example per label, LUKE obtains a 0 macro F1 score whereas ZERO obtains 0.5507 or a 0.2766 improvement from the zero-shot learning setting.

\subsection{Limitations}

With an average macro F1 score of 0.2323 for zero-shot-learning, the performance might not yet be good enough for practical applications. For example, for patient note de-identification, this performance would not be acceptable since it would fail to remove a large amount of the protected health information.

Furthermore, the small size of the CrossNER dataset may result in both greater uncertainty of the results and lower performance of the trained model. Other large datasets such as CoNLL-2003 for NER exist, but they are limited to a single domain. These problems could be solved if larger datasets with more samples per label and more domains existed.

\section{Conclusion}

We propose a zero-shot learning NER approach with external knowledge. In contrast to traditional NER models like LUKE, our model utilises dot products of label entity features and token features to classify, allowing it to perform inference without retraining on the target domain. Experiments show that our model outperforms LUKE in zero-shot and few-shot learning; meanwhile, it also gets comparable results with LUKE in supervised in-domain recognition.

Future work may explore more recent SOTA pre-trained language models and word embeddings. Furthermore, a solution to the limited size of the CrossNER dataset would be to artificially create a larger cross-domain dataset. One could, for example, use a dataset such as CoNLL-2003 which is based on the domain of news for training and then test the performance on another large dataset such as AnEM which is based on the domain of anatomy. Thus the zero-shot learning method is preserved while allowing for much larger training datasets and hence better generalization.

\newpage

\section{Broader Impact}

Improvements in cross-domain NER has a large number of beneficial applications for society, namely the ones that benefit from being able to perform NER in a domain where little to no training data is available. However, we understand that it also has potentially harmful consequences.

In case of failure of the NER system, it can have severe implications depending on the domain and use case. E.g.\ for the application of patient note de-identification, it would mean that protected health information would be kept in the medical notes that were to be distributed to a larger population, severely harming the anonymity of the individuals who happened to be kept in the medical notes.

ZERO is also likely display certain biases towards different genders, ethnicity etc. This is largely due to the fact that our proposed system utilizes word embeddings for the external knowledge, which have been shown to include biases \cite{Bolukbasi2016}. For a dataset like CrossNER, if e.g.\ the embedding of `scientist' is closer to the embeddings of male or western-sounding names than to those of female or black-sounding names, then it would mean minorities would more often be missed by the system, enforcing stereotypes and possibly harming publicity and career aspects of the individuals affected. There are known methods for minimizing biases in word embeddings and these should be applied before deploying ZERO in a production scenario \cite{Bolukbasi2016, ananya2019genderquant}.

\bibliography{anthology,custom}
\bibliographystyle{acl_natbib}

\end{document}